# AI-Driven Carbon Monitoring: Transformer-Based Reconstruction of Atmospheric $CO_2$ in Canadian Poultry Regions

Padmanabhan Jagannathan Prajesh[1,2], Kaliaperumal Ragunath[3], Miriam Gordon[1], Bruce Rathgeber[1], and Suresh Neethirajan[1,4*]

- [1]Faculty of Agriculture, Dalhousie University, Truro, Nova Scotia B2N 5E3, Canada.
- [2]Department of Remote Sensing and GIS, Tamil Nadu Agricultural University, Coimbatore 641003, India.
- [3]Center for Water and Geospatial Studies, Tamil Nadu Agricultural University, Coimbatore 641003, India.
- [4]Faculty of Computer Science, Dalhousie University, Halifax, Nova Scotia B2N 5E3, Canada.

**Corresponding author:** Suresh Neethirajan[1,4], sneethir@gmail.com

**Abstract**

Accurate mapping of column-averaged $CO_2$ ($XCO_2$) over agricultural landscapes is essential for guiding emission mitigation strategies. We present a Spatiotemporal Vision Transformer with Wavelets (ST-ViWT) framework that reconstructs continuous, uncertainty-quantified $XCO_2$ fields from OCO-2 across southern Canada, emphasizing poultry-intensive regions. The model fuses wavelet time–frequency representations with transformer attention over meteorology, vegetation indices, topography, and land cover. On 2024 OCO-2 data, ST-ViWT attains $R^2$=0.984 and RMSE=0.468 ppm; 92.3% of gap-filled predictions lie within ±1 ppm. Independent validation with TCCON shows robust generalization (bias −0.14 ppm; r=0.928), including faithful reproduction of the late-summer drawdown. Spatial analysis across 14 poultry regions reveals a moderate positive association between facility density and $XCO_2$ (r=0.43); high-density areas exhibit larger seasonal amplitudes (9.57 ppm) and enhanced summer variability. Compared with conventional interpolation and standard machine-learning baselines, ST-ViWT yields seamless 0.25° $CO_2$ surfaces with explicit uncertainties, enabling year-round coverage despite sparse observations. The approach supports integration of satellite constraints with national inventories and precision livestock platforms to benchmark emissions, refine region-specific factors, and verify interventions. Importantly, Transformer-based Earth observation enables scalable, transparent, spatially explicit carbon accounting, hotspot prioritization, and policy-relevant mitigation assessment.

## 1. Introduction

Accurate, spatially resolved monitoring of column-averaged dry air $CO_2$ ($XCO_2$) is increasingly critical for constraining greenhouse gas (GHG) budgets, evaluating mitigation strategies, and informing climate-smart agricultural management (Mustafa et al., 2021). While global attention has focused largely on energy and transportation sectors, agriculture sector contributes substantially to anthropogenic $CO_2$ fluxes, accounting for approximately 11.9% of global GHG emissions via direct and indirect pathways (Hur et al., 2024). Although poultry production is often considered a minor emitter relative to ruminants and crop production, life-cycle assessments indicate significant carbon footprints, with emission intensities averaging 1.26 kg $CO_2e$ from feed, 0.55 kg from transport, 0.28 kg $N_2O$ from manure, and 0.26 kg from farm energy per kilogram of product (Oke et al., 2024). As global poultry demand continues to grow, accurately quantifying its full carbon budget is essential for developing effective, sector-specific mitigation strategies.

Bottom-up GHG inventories, derived from activity data and emission factors, often obscure fine-scale spatial heterogeneity and temporal variability (Kivimäki et al., 2025). Poultry systems exemplify this complexity, combining feed production, housing, manure storage, and energy use, all contributing unevenly to $CO_2$ fluxes (Grossi et al., 2019; U.S. EPA, 2021). Regional contrasts in Canada further amplify these differences as Ontario and Quebec host dense broiler operations; British Columbia's Fraser Valley uses distinct manure management practices; and Prairie provinces face elevated heating demands due to continental climates (Statistics Canada, 2022). Such structural variability produces unique emission profiles through differences in feed sourcing, climate-control energy, and production management, suggesting that conventional bottom-up inventories may underestimate localized emissions, highlighting the need for AI-driven sensor innovations (Neethirajan, 2023) and spatial benchmarking (Kuhlmann et al., 2021) for comprehensive atmospheric monitoring.

Advances in spaceborne $XCO_2$ observations from GOSAT/GOSAT-2, OCO-2/OCO-3, and CarbonSat now provide near-global coverage with bias-corrected retrievals consistent with measurements from the ground-based validation networks, exhibiting mean biases of 0.02-0.25 ppm and standard deviations (SD) around 2 ppm, with precision typically less than 0.8 ppm (Kulawik et al., 2019). However, narrow swaths, clouds, aerosols, and viewing geometry create structured data gaps (Hu et al., 2024), challenging emission attribution in heterogeneous landscapes (Watanabe et al., 2023). Machine learning methods, including ensemble models and neural networks, enable gap-filling and continuous $XCO_2$ reconstruction using land-atmosphere predictors, achieving $R^2$ of 0.95–0.98 and RMSE of 0.58–0.91 ppm (He et al., 2024; Alcibahy et al., 2025). While convolutional and LSTM networks improve spatiotemporal modeling, long-range transport and seasonal variability remain challenging (Hu et al., 2024). Multi-modal fusion and transformer-based approaches reduce monthly bias by up to 71.5% and extend temporal coverage from 14 to 21.5 days, enhancing the completeness and reliability of atmospheric $CO_2$ fields (Dumont Le Brazidec et al., 2024).

Deep learning has transformed spatiotemporal Earth observation, particularly via transformer-based architectures leveraging self-attention to capture global dependencies across space and time (Dosovitskiy et al., 2021). GeoFormer has demonstrated high performance, correlating 0.94–0.95 with TCCON GHG measurements (Khirwar and Narang, 2024). Spatiotemporal Vision Transformers (ST-ViTs) further enhance modeling by integrating multi-scale spatial features and extended temporal dynamics, mitigating vanishing gradient issues through parallelized self-attention (Arnab et al., 2021). ST-ViTs can incorporate auxiliary datasets with feature-specific weighting, providing ecological and biophysical context that strengthens predictions and enables reconstruction of missing $XCO_2$ in heterogeneous agricultural landscapes (Jung et al., 2020; Ruehr et al., 2023). Domain-specific transformers, such as MethaneMapper for hyperspectral methane detection (Kumar et al., 2023) and GasFormer for livestock emission segmentation (Sarker et al., 2024), exemplify the growing potential of transformers for high-resolution atmospheric monitoring.

Building on these advances, this study addresses the gap in gap-filled, uncertainty-quantified $XCO_2$ data over poultry-intensive regions in Canada, where conventional inventories rely on generalized emission factors that obscure spatial heterogeneity and seasonal dynamics. We introduce a Spatiotemporal Vision Transformer with Wavelets (ST-ViWT) to reconstruct continuous $XCO_2$ fields, capturing non-stationary seasonal dynamics through wavelet-based time-frequency representations and integrating spectral tokens with meteorological, and topographical embeddings. The model produces a seamless 0.25° $XCO_2$ surface with per-grid uncertainty, preserving spatiotemporal variability. This approach has direct policy relevance, as Canada's GHG accounting continues to rely on Tier 1/2 emission factors (Environment and Climate Change Canada, 2025). Observation-constrained fields independently validate bottom-up emissions estimates, inform region-specific factors (Watanabe et al., 2023), and when coupled with livestock data, enable benchmarking and transparency for offsets and sustainability reporting.

## 2. Methods and Materials

### 2.1 Study Area and Data Sources

The study focused on southern Canada, characterized by intensive agriculture and diverse physiographic conditions. The domain, delineated using FAO GAUL 2024 Level 2 administrative boundaries (FAO, 2024), spans 41.7°–83.1° N and 141.0°–52.6° W, encompassing spatial heterogeneity in poultry operation density across distinct climatic and agroecological zones (Supplementary File S1).

#### 2.1.1 Primary $XCO_2$ Dataset

The primary dataset consisted of Level 2 Orbiting Carbon Observatory-2 (OCO-2) $XCO_2$ retrievals for 2024, providing high-precision column-averaged $CO_2$ mole fractions (Eldering et al., 2017) with global bias corrections $\leq$0.2 ppm relative to Total Carbon Column Observing Network (TCCON) measurements (Jacob et al., 2024). Initial data coverage (Fig. 1) exhibited the characteristic irregular spatiotemporal sampling of polar-orbiting satellites, with gaps induced by cloud cover, aerosols, and orbital constraints.

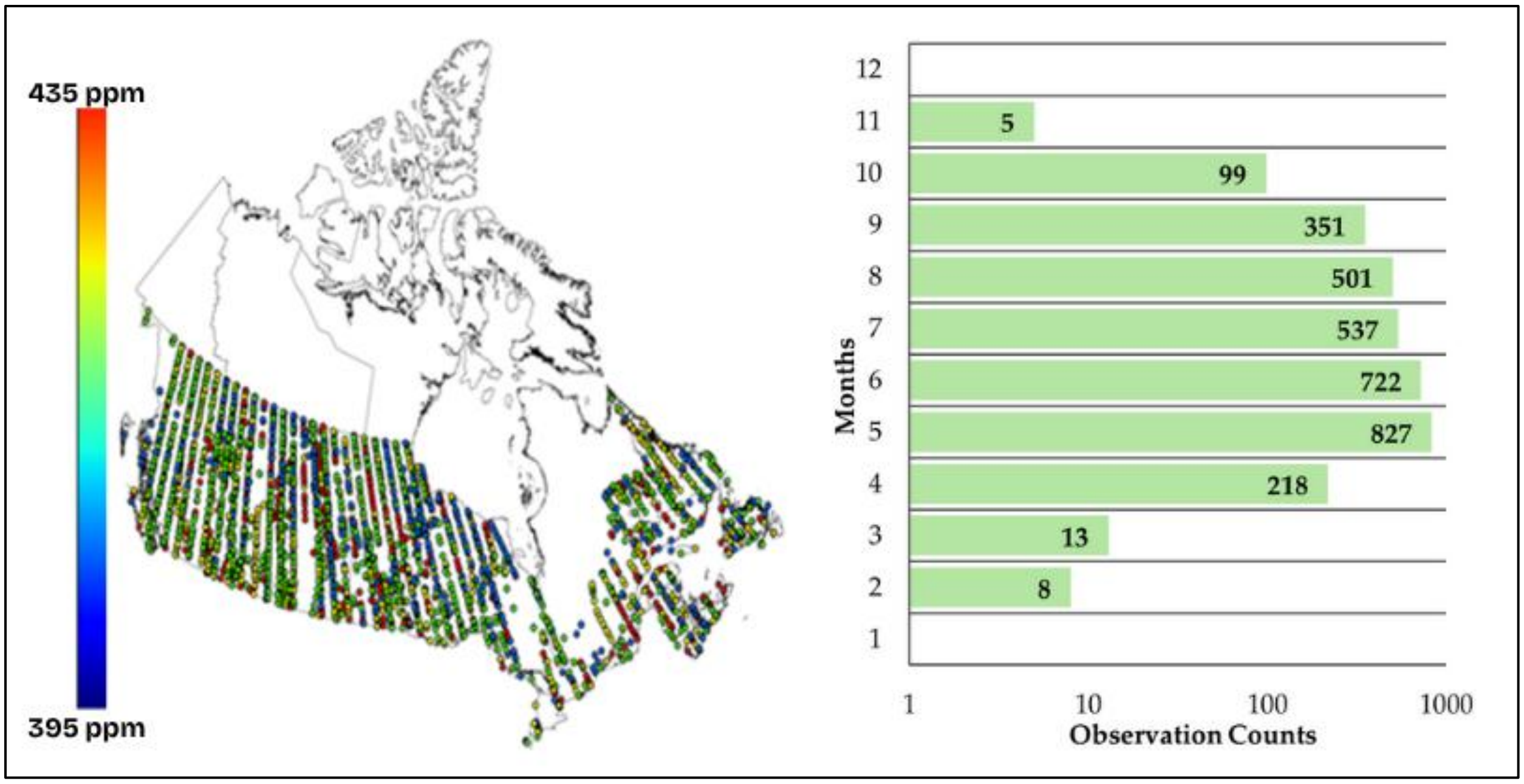

**Fig. 1. Spatial Distribution and Monthly OCO-2 $XCO_2$ Observations In 2024.**

### 2.1.2 Auxiliary Geospatial Datasets

Multiple geospatial auxiliary datasets were integrated to enhance $XCO_2$ reconstruction and provide environmental context, selected for relevance, availability, and temporal alignment with OCO-2 (Table 1). Static variables (elevation, land cover) used nearest-neighbor matching, while dynamic variables (vegetation indices) employed temporal averaging/interpolation. Quality control included range checks, inter-variable consistency, spatial continuity, and temporal stability assessments to ensure data reliability and coherence.

**Table 1. Auxiliary Geospatial Datasets and Their Applications in $XCO_2$ Reconstruction**

| Dataset | Variables | Role in $XCO_2$ Reconstruction |
|---|---|---|
| ERA5-Land Daily Aggregated | Temperature, precipitation, wind, pressure. | Controls photosynthesis/respiration rates, atmospheric mixing, and transport patterns. |
| MODIS MOD13A1 | NDVI, EVI. | Indicates vegetation $CO_2$ uptake; correlates with GPP and seasonal $CO_2$ variations. |
| SRTM 1 Arc-Second | Elevation, Slope (Derived). | Affects column corrections, vegetation distribution, and regional $CO_2$ patterns. |
| ESA WorldCover 10 m | Land cover, Agricultural areas, | Differentiates anthropogenic sources from natural sinks. |
| AAFC Crop Inventory 30 m | Agricultural intensity. | Links feed crop areas to poultry density and associated $CO_2$ emissions. |

## 2.2 Grid Aggregation and Workflow

A uniform 0.25° × 0.25° spatial grid was adopted as the reference framework for data integration, balancing computational efficiency with sufficient resolution to capture adequate representation of atmospheric $CO_2$ transport processes and align with atmospheric transport model scales of 100-500 km (Gadhavi et al., 2025). OCO-2 observations were spatially binned via nearest-neighbor assignment, with each cell's mean, median, standard deviation, minimum, maximum, and observation count computed. Temporal variability was retained using harmonic analysis of monthly values (sine/cosine transforms) and day-of-year metrics to support seasonal and phenological modeling. The complete pipeline is illustrated in Fig. 2, showing the flow from input data through preprocessing, wavelet transformation, embedding, transformer processing, and final reconstruction.

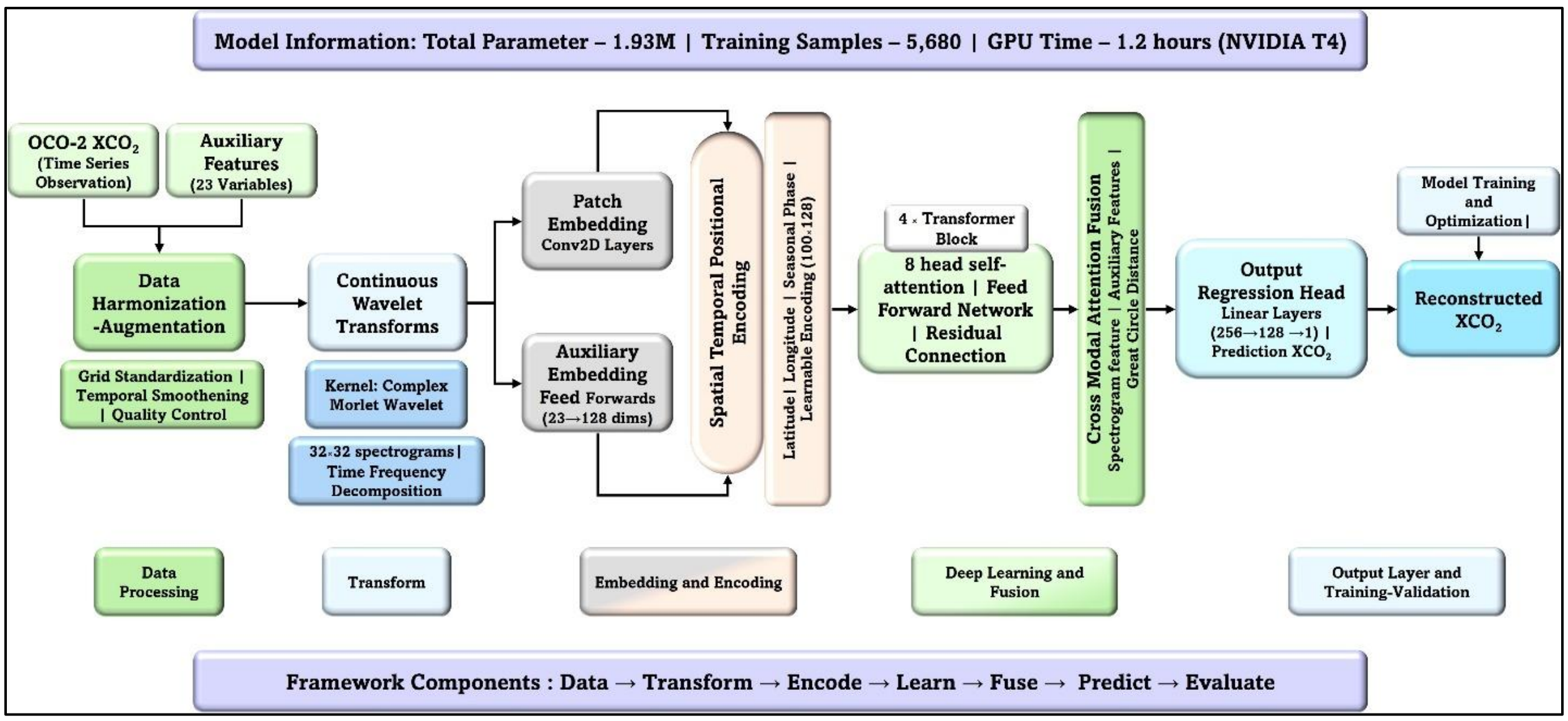

**Fig. 2 Model Architecture Overview with Complete Pipeline From OCO-2 $XCO_2$ Input Through Reconstruction**

### 2.3 Spatiotemporal Feature Design

An advanced feature engineering framework was developed to represent the spatiotemporal complexity of atmospheric $CO_2$ dynamics, integrating temporal harmonics, spatial interaction terms, distance-based encodings, and seasonal aggregations to capture both localized variability and large-scale atmospheric drivers. The corresponding formulations and inferences are provided in Supplementary File S2.

Temporal harmonics were incorporated using sine and cosine transformations of annual and semi-annual cycles to capture continuous intra-annual variability (Miller and Michalak, 2020). Spatial heterogeneity was modeled through higher-order geographic interaction terms derived from latitude and longitude, including quadratic, cross, and centroid-referenced distance encodings that represent meridional and zonal $CO_2$ gradients (Gurney et al., 2002). A distance-decay weighting function with a 5° spatial scale emphasized proximity to Canada's geographic centroid while preserving large-scale atmospheric correlations (Dadheech et al., 2024). Cross-interaction terms between spatial and temporal dimensions captured latitude-dependent seasonal amplitudes and longitudinal phase shifts influenced by oceanic-continental contrasts (He et al., 2024). Seasonal aggregation indicators employed weighted functions emphasizing vegetation-active summer months, reflecting their dominant influence on annual $CO_2$ budgets.

### 2.4 Spatiotemporal Vision Transformer with Wavelets (ST-ViWT) Framework

The ST-ViWT framework (Table 2) integrates continuous wavelet transforms (CWT) with a transformer architecture, enabling cross-modal attention with auxiliary geospatial features. This design overcomes limitations of traditional Fourier methods in capturing non-stationary temporal patterns characteristic of atmospheric $CO_2$ dynamics (Mustafa et al., 2021; Wang et al., 2024).

**Table 2. ST-ViWT Architecture Components**

| Component | Specification | Implementation Details |
|---|---|---|
| Patch Embedding | Conv2D (1→128, 4×4, stride=4) | Converts 32×32 spectrograms to tokens |
| Auxiliary Embedding | Linear (23→128) + ReLU + Dropout | Processes auxiliary features |
| Positional Encoding | Learnable (1×100×128) | Encodes spatial relationships |
| Transformer Blocks | 4×8 attention heads | Multi-head self-attention |
| Output Head | LayerNorm + Linear (256→128→1) | Regression prediction |

Atmospheric $XCO_2$ time series were transformed into time-frequency spectrograms using CWT with complex Morlet kernels, producing normalized 8-bit intensity inputs for the network (Mustafa et al., 2021; Reichert et al., 2024). Grid-averaged observations were augmented with climatological sinusoidal cycles aligned to Northern Hemisphere seasonal minima where seasonal amplitude is 3 ppm and phase offset recorded is $(-\frac{\pi}{2})$ to improve temporal continuity in under-sampled regions (Bastos et al., 2020). A multi-head self-attention module with adaptive scaling $\alpha_{scale}$ was introduced to modulate attention weights by wavelet scale, enhancing sensitivity to multi-frequency temporal structures. Learnable latitude-longitude-season encoding was implemented to preserve spatial proximity and temporal phase relationships within the transformer's latent space (Mao et al., 2025). Auxiliary variables (meteorology, topography, vegetation) were embedded via a feed-forward network and fused with spectrogrammatic representations using distance-weighted encoding:

$$F_{spatial}\,(i,j) \;=\; e^{-\gamma d_{ij}} \,\cdot\, F_{aux,j}$$

Where, $d_{ij}$ is the great-circle distance between cells $i$ and $j$, and $\gamma$ is a learnable decay parameter. This ensures geographically smooth fusion consistent with atmospheric transport behavior. ST-ViWT was trained for spatiotemporal $XCO_2$ reconstruction using an 80/20 spatially stratified train-test split to minimize geographic bias. Reproducibility was ensured via a fixed seed (42) for data splitting, weight initialization, and dropout. Batches of 32×32 spectrogram tensors with 23 auxiliary features were optimized using AdamW (lr = $1\times10^{-4}$, weight decay = $1\times10^{-5}$) with cosine annealing and MSE loss, converging in ~52 epochs. Training on 5,680 OCO-2 samples required ~1.2 GPU hours on an NVIDIA T4. With 1.96 M parameters, the model balances high predictive performance with computational efficiency. The corresponding training details are summarized in the Supplementary File S3.

### 2.5 Poultry Density and $XCO_2$ Estimation

Poultry facility density ($\rho$) was computed as the number of facilities per region, normalized by geodetically corrected area to account for Earth's curvature:

$$\rho = \frac{N_{facilities}}{A_{region} \times cos(lat_{mean})}$$

Where, $N_{facilities}$ is the facility count, $A_{region}$ is the nominal area, and $lat_{mean}$ is the region's mean latitude. Facility counts were derived via spatial intersection, and densities expressed in facilities per $km^2$. Density values were categorized into tertile-based classes: ($<33^{rd}$ percentile), medium ($33^{rd}$-$67^{th}$ percentile), and high ($>667^{th}$ percentile) enabling standardized inter-regional comparisons across 14 distinct poultry production regions in southern Canada.

### 2.6 TCCON-Based Multi-Scale $XCO_2$ Validation

The East Trout Lake TCCON station (54.35°N, 104.99°W - https://tccondata.org/ ) provides high-precision $XCO_2$ measurements, enabling independent validation of reconstructed fields (Wunch et al., 2025). In this study, ST-ViWT predictions were benchmarked against 3,38,406 TCCON observations using a multi-radius spatial matching framework with radii of 0.5°, 1.5°, and 2.5° to evaluate scale-dependent accuracy. This approach quantifies model performance in both data-dense and sparse regions (Gadhavi et al., 2025), addressing footprint-to-grid resolution mismatches and providing a transferable protocol for validating satellite-derived and reconstructed $XCO_2$ in heterogeneous landscapes (Das et al., 2025).

## 3. Results

### 3.1 Model Development and Training Performance

#### 3.1.1 Training Dynamics and Convergence Behavior of ST-ViWT Model

The ST-ViWT model demonstrated stable convergence over 60 training epochs, with both training and validation losses decreasing monotonically (Fig. 3). Loss values were reported in parts per million squared ($ppm^2$), reflecting the use of the mean squared error (MSE) as the objective function, where the squared unit indicates the average squared deviation between predicted and true $XCO_2$ concentrations. Initial validation loss of 0.360 $ppm^2$ declined by 50% within the first 10 epochs, while training loss dropped from 0.522 $ppm^2$ to 0.189 $ppm^2$, demonstrating efficient early learning. Between epochs 10 and 35, both losses continued to decrease steadily. validation loss fell from 0.140 $ppm^2$ to 0.045 $ppm^2$, and training loss from 0.169 $ppm^2$ to 0.058 $ppm^2$, indicating minimal overfitting and consistent improvement in predictive accuracy.

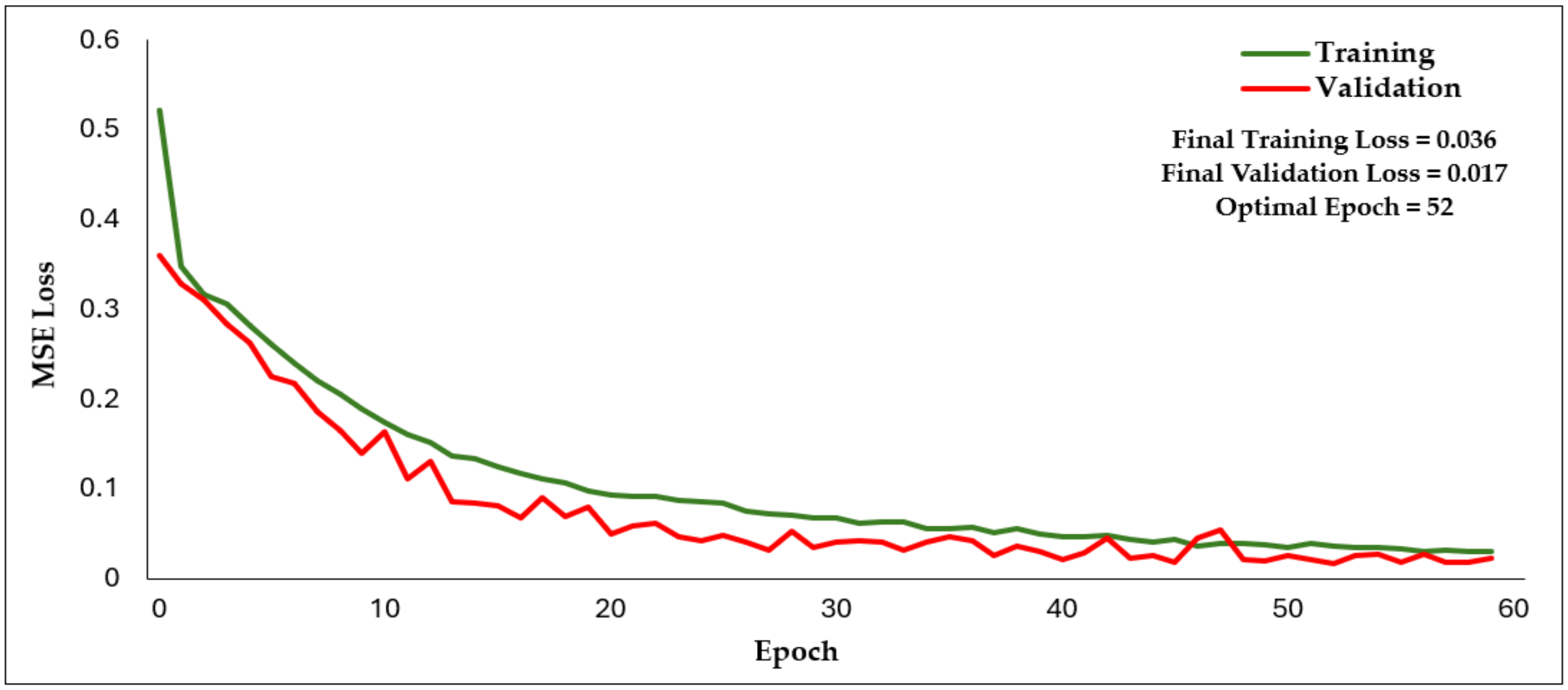

**Fig. 3 Training and Validation Loss of ST-ViWT Model Over 60 Epochs**

A consistent gap between training and validation losses indicated effective regularization, with no significant oscillations or divergence. The lowest validation loss of 0.0169 ppm² was achieved at epoch 52, representing a 95.3% reduction from the initial value, while training loss stabilized at 0.0361 ppm². The 2:1 ratio between training and validation losses persisted throughout, reflecting balanced model complexity. During the final convergence phase (epochs 45-60), validation loss improvements were marginal (<0.005 ppm² per epoch), indicating approach to an asymptotic minimum. No secondary minima or plateaus were observed, demonstrating that the optimization strategy effectively navigated the loss landscape.

### 3.1.2 Model Accuracy and Statistical Performance Validation

The ST-ViWT model achieved high predictive accuracy for satellite-based $XCO_2$ reconstruction (Fig. 4), with a validation $R^2$ of 0.984, capturing 98.4% of atmospheric $CO_2$ variance. Performance was consistent across datasets, with training and test $R^2$ of 0.987 and 0.982, respectively, and minimal 0.3% degradation, indicating appropriate complexity without overfitting.

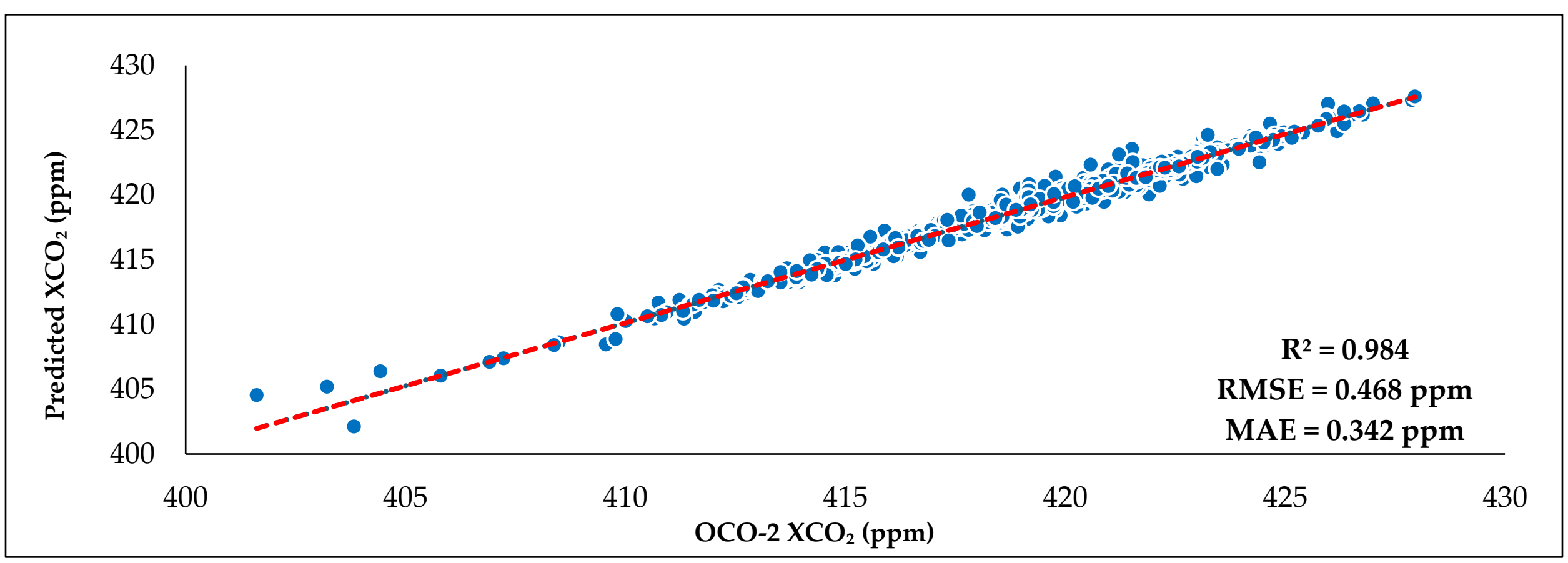

**Fig. 4 Comparison of ST-ViWT $XCO_2$ Predicted vs Observed**

The Mean Absolute Error (MAE) increased slightly from training (0.318 ppm) to validation (0.342 ppm) and test (0.356 ppm), while RMSE remained low (0.445-0.479 ppm), corresponding to less than 0.12% relative error. Systematic bias was minimal with mean bias ranging from -0.098 to -0.145 ppm), with median bias smaller (-0.067 to -0.102 ppm), confirming limited outlier influence. Adjusted $R^2$ values (0.981-0.986) closely matched standard $R^2$, and bias standard deviations (0.441-0.471 ppm) approximated RMSE, indicating predominantly random errors. Prediction distributions showed near-normal behavior, with skewness (0.198-0.234) and kurtosis (3.289-3.456) supporting well-calibrated, reliable uncertainty estimates.

### 3.1.3 Error Distribution and Model Reliability

The ST-ViWT model exhibited a near-Gaussian prediction error distribution (Table 3), indicating well-calibrated and statistically robust performance. The mean residual was 0.000 ± 0.468 ppm, reflecting unbiased predictions across the full $XCO_2$ range. Residuals showed slight negative skewness (-0.118), suggesting a minor underestimation at higher concentrations, while kurtosis (2.94) aligned with a normal distribution. Error variance remained homoscedastic, confirming stable performance regardless of absolute concentration. The maximum absolute error was 2.917 ppm, with only 0.26% of predictions exceeding 2 ppm; 99.74% fell within ±2.5 ppm.

**Table 3. Error Distribution and Model Reliability Metrics**

| Error Metric | Value | Interpretation |
|---|---|---|
| **Central Tendency** | | |
| Mean Residual | 0.000 ± 0.468 ppm | Unbiased predictions |
| Median Absolute Error | 0.289 ppm | Typical prediction error |
| **Distribution Shape** | | |
| Residual Skewness | -0.118 | Nearly symmetric |
| Residual Kurtosis | 2.94 | Normal distribution |
| **Reliability** | | |
| 95$^{th}$ Percentile Error | 1.234 ppm | High confidence bound |
| 99$^{th}$ Percentile Error | 2.456 ppm | Extreme case bound |
| Maximum Absolute Error | 2.917 ppm | Worst-case scenario |
| **Coverage Statistics** | | |
| Predictions within ±1 ppm | 92.3 | Excellent |
| Predictions within ±2 ppm | 99.0 | Outstanding |
| Predictions within ±2.5 ppm | 99.74 | Near perfect |
| **Uncertainty Quantification** | | |
| 95% Confidence Interval | ±0.917 ppm | |
| Prediction Reliability Index | 0.976 | |

The 95% confidence interval (±0.917 ppm) captured the majority of errors, providing a robust uncertainty bound for scientific and policy applications. Larger residuals were primarily located in coastal and complex topographic regions, reflecting structural satellite data limitations rather than model instability. The histogram of residuals (Fig. 5) confirmed a unimodal, near-symmetric distribution without significant outliers, supporting the assumption of normally distributed errors essential for reliable uncertainty quantification.

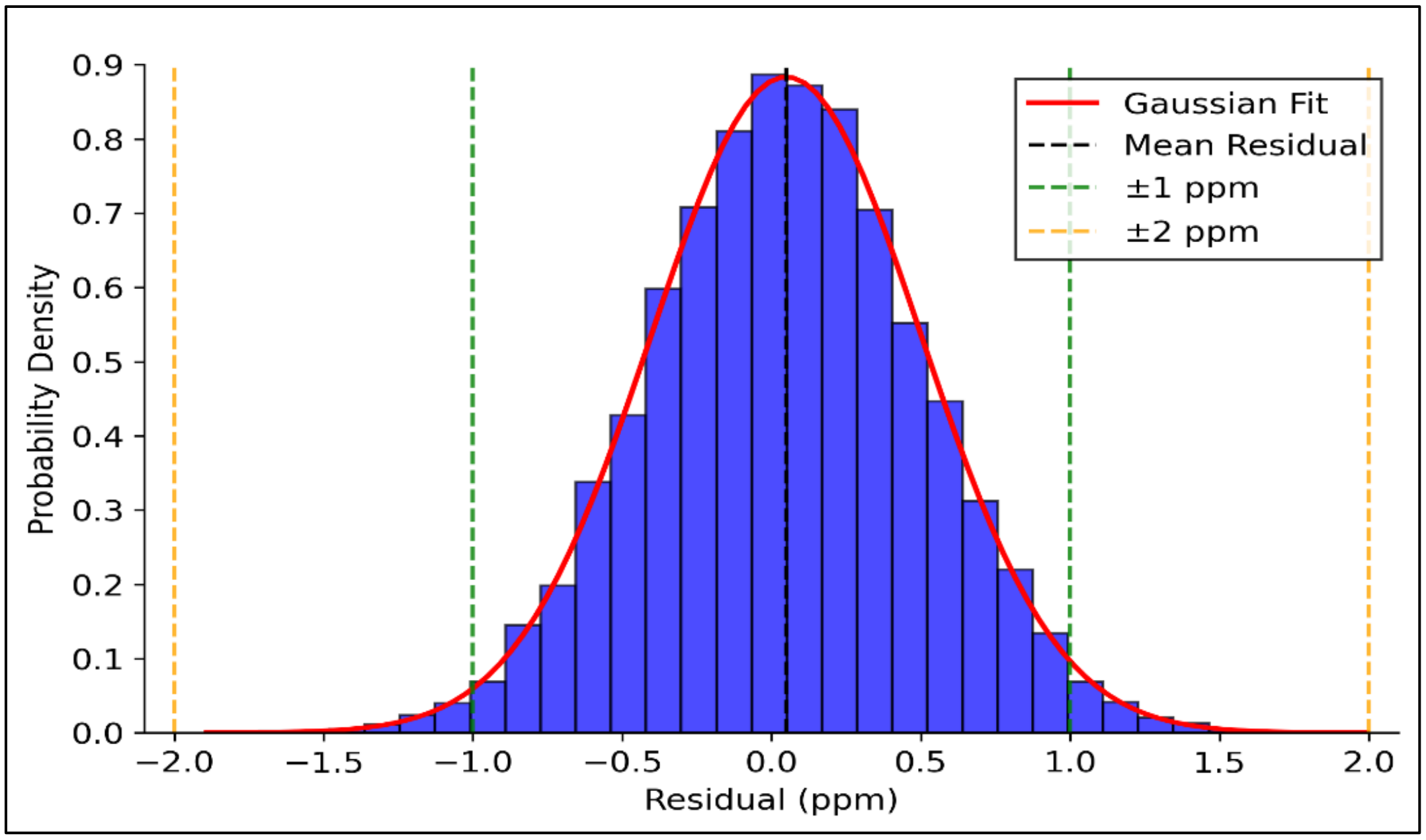

**Fig. 5 Residual Histogram of ST-ViWT Model Prediction**

### 3.2 $XCO_2$ Dataset Characteristics and Coverage Analysis

#### 3.2.1 Temporal Coverage and Observation Density

The temporal distribution of the $XCO_2$ dataset (Table 4) exhibits pronounced seasonal variability, reflecting the latitudinal and diurnal constraints imposed by the OCO-2 satellite's sun-synchronous orbit.

**Table 4. Seasonal Distribution of $XCO_2$ Observations**

| Season | Months | Observation Percentage (%) |
|---|---|---|
| Summer | June-August | 56.5% |
| Spring | March-May | 26.8% |
| Fall | September-November | 14.8% |
| Winter | December-February | 1.9% |
| **Growing Season Total** | **May-September** | **77.4%** |

Observations were strongly biased toward summer season (June-August), which accounted for 56.5% of the dataset. Spring (March-May) and fall (September-November) contributed 26.8% and 14.8%, respectively, while winter (December-February) was markedly underrepresented at only 1.9%. Notably, 77.4% of all observations occurred within the growing season from May to September, a period characterized by enhanced retrieval reliability due to reduced cloud cover and higher solar zenith angles at high northern latitudes. This pronounced seasonal skew necessitated the implementation of temporal interpolation and data augmentation strategies to support year-round model training and inference, thereby mitigating the risk of overfitting to warm-season conditions.

#### 3.2.2 Spatial Coverage of $XCO_2$ Datasets

The observed and reconstructed $XCO_2$ datasets (Supplementary File S4) provide spatially comprehensive satellite-based coverage across Canada, encompassing diverse regions, ecosystems, and temporal intervals. Direct OCO-2 observations ranging from 392.49 ppm to 438.02 ppm with a SD of 5.44 ppm and an average sampling density of 0.97 observations per grid cell. Model-based reconstructions complemented the remaining 3.4% of the grid at a lower sampling density (0.03 per cell), while demonstrating strong agreement with observations (mean = 419.12 ppm; uncertainty = 0.58 ppm). Integration yielded 100% spatial coverage (Fig. 6) with a harmonized mean of 418.18 ppm (SD 5.45 ppm) and reduced uncertainty of 0.36 ppm. This quality-controlled, high-resolution dataset supports robust regional-to-continental analyses, including carbon flux inversions, atmospheric transport modeling, and greenhouse gas climatology.

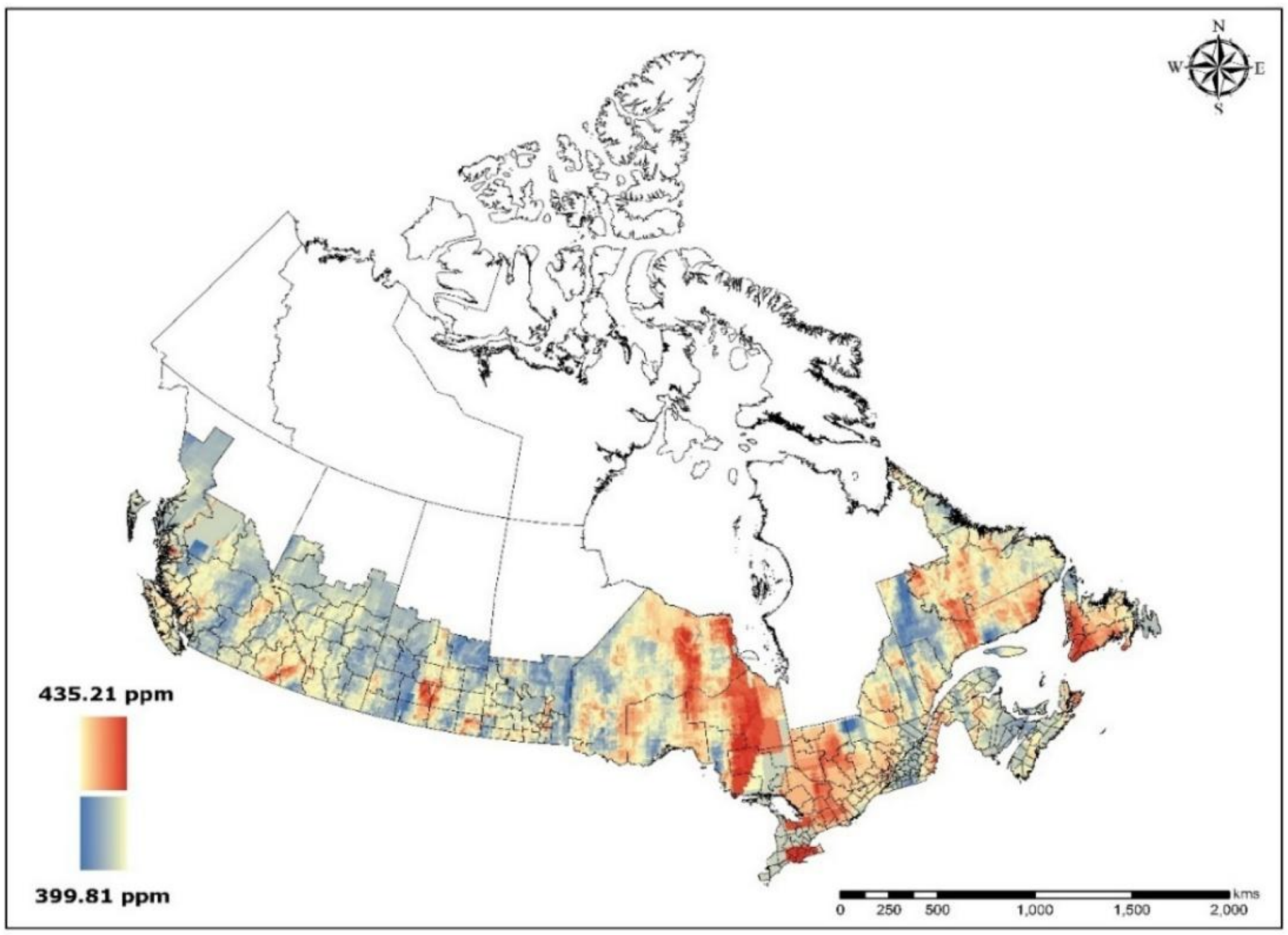

**Fig. 6 Reconstructed $XCO_2$ (ppm) Spatial Distribution in Southern Canada, 2024**

### 3.3.3 Temporal Dynamics of Reconstructed $XCO_2$ Concentration

Monthly (Fig. 7) and seasonal (Fig. 8) analyses of reconstructed $XCO_2$ across Canada reveal a pronounced and coherent seasonal cycle consistent with Northern Hemisphere carbon dynamics. Monthly means follow a sinusoidal trend, peaking in spring and reaching a minimum in late summer, driven by biospheric uptake and atmospheric accumulation. Winter concentrations remain stable, with January averaging 420.45 ± 2.20 ppm and February 421.32 ± 1.62 ppm, reflecting low terrestrial exchange and stable meteorology. The spring transition begins in March (422.59 ± 2.10 ppm) and peaks in April (423.89 ± 1.74 ppm), representing maximum atmospheric $CO_2$ before widespread photosynthetic drawdown. During the growing season, $XCO_2$ declines sharply from 422.63 ± 1.84 ppm in May to a minimum of 414.97 ± 3.01 ppm in August. August shows the highest variability, with standard deviations nearly double those of winter, reflecting heterogeneous vegetation growth, regional ecosystem diversity, and differing phenological timing. Autumn recovery spans September (416.40 ± 2.41 ppm) to November (420.89 ± 1.94 ppm), returning to near the winter baseline by December (422.05 ± 1.66 ppm). The overall seasonal amplitude of 8.92 ppm (April–August) aligns with northern mid-latitude estimates, validating the model's representation of continental carbon dynamics.

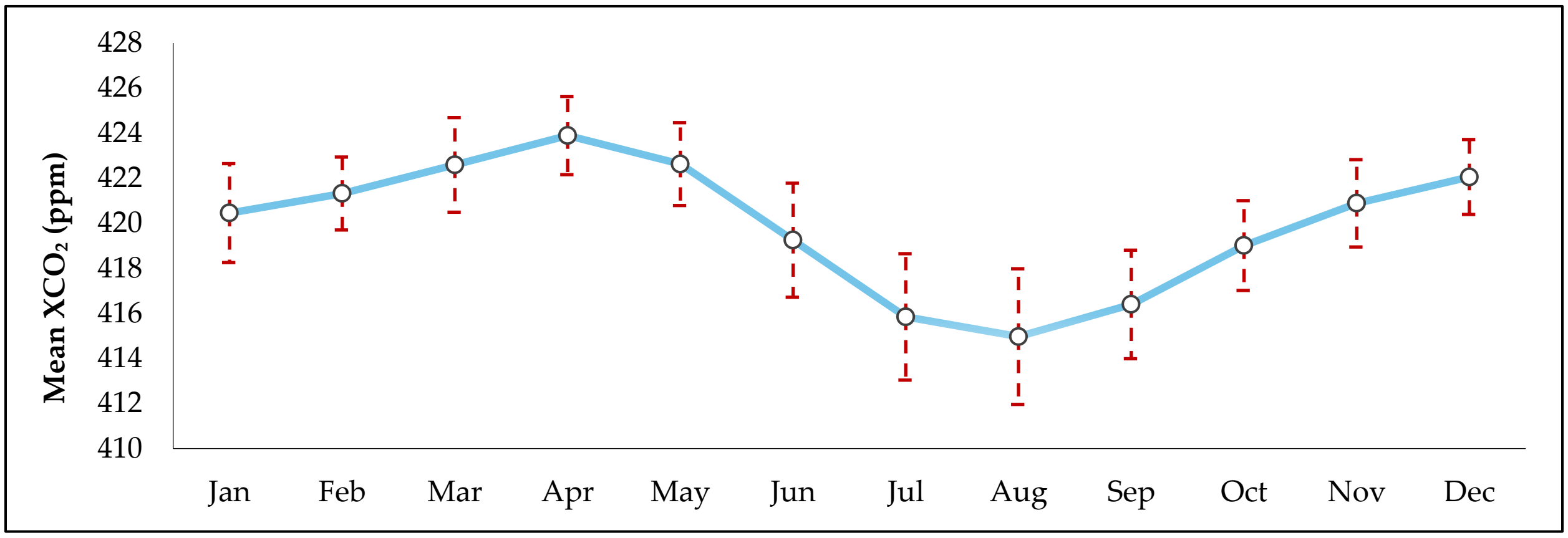

**Fig. 7 Monthly Mean of $XCO_2$ Concertation with standard deviation**

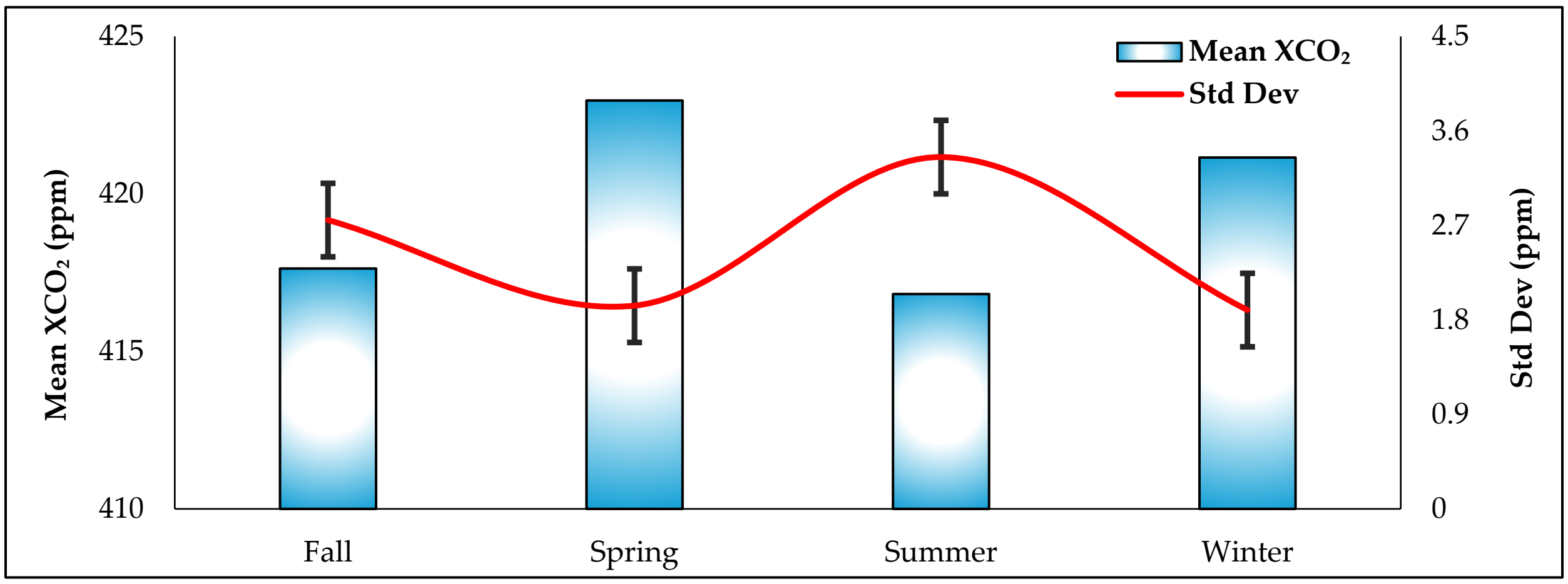

**Fig. 8 Mean Seasonal $XCO_2$ Concertation with standard deviation**

Seasonal aggregation further highlights ecosystem influence: winter averages 420.90 ± 2.18 ppm (IQR = 2.89 ppm) with minimal biological activity; spring 422.50 ± 1.98 ppm, moderately skewed toward higher values; summer 417.36 ± 2.71 ppm with highest variability (IQR = 3.84 ppm); and fall 417.38 ± 2.35 ppm, maintaining low concentrations and reduced variability. Seasonal coefficients of variation range from 0.52% (winter) to 0.65% (summer), highlighting the contrast between stable winter conditions and dynamic summer processes (spatial observation metrics are in Supplementary File S4).

### 3.3 Reconstructed $XCO_2$ Variability Across Canadian Poultry Regions

#### 3.3.1 Correlation Between Poultry Facility Density and Atmospheric $XCO_2$

The analysis of atmospheric $XCO_2$ concentrations across 14 major Canadian poultry-producing regions revealed a moderate positive correlation between poultry facility density and annual average $XCO_2$ ($r = 0.434$, $p < 0.05$). Poultry facility density explained approximately 19% of the total variance in $XCO_2$ concentrations, while the remaining 81% can be attributable to meteorological, topographic, and other environmental influences. Annual averages demonstrated a clear density-dependent gradient (Supplementary File S5). High-density regions (≥0.0015 facilities/km²) recorded the highest $XCO_2$ (419.15 ± 1.22 ppm), followed by medium-density regions (0.00075-0.00149 facilities/km²; 419.07 ± 1.07 ppm), and low-density regions (<0.00075 facilities/km²; 418.26 ± 1.41 ppm).

Notable anomalies highlighted environmental modulation of poultry-driven signals. The Nova Scotia Annapolis Valley, despite high facility density (0.0023 facilities/km²), exhibited the second-lowest mean $XCO_2$ (417.21 ppm), reflecting strong atmospheric mixing from maritime winds. In contrast, Manitoba Eastern displayed elevated $XCO_2$ (420.42 ppm) despite medium density (0.0012 facilities/km²), likely due to reduced mixing in continental interior settings. The BC Fraser Valley, with the highest density (0.0176 facilities/km²), also showed elevated $XCO_2$ (420.74 ppm), consistent with valley topographic trapping of emissions.

### 3.3.2 Spatial and Density-Dependent Variations of $XCO_2$ in Poultry Regions

Aggregated distributions (Fig 9) revealed overlapping $XCO_2$ ranges among density categories. Low-density regions showed the greatest variability ($\sigma = 1.41$ ppm), while medium-density areas were most stable ($\sigma = 1.07$ ppm), suggesting low-density systems are more susceptible to heterogeneous meteorological influences. Across all regions, the total $XCO_2$ range of 4.95 ppm (~1.2% of baseline atmospheric $CO_2$) confirms the detectability of poultry-related signals within broader environmental variability. Across all regions, the total $XCO_2$ range of 4.95 ppm corresponds to ~1.2% of baseline atmospheric $CO_2$, confirms the detectability of poultry-related signals within broader environmental variability.

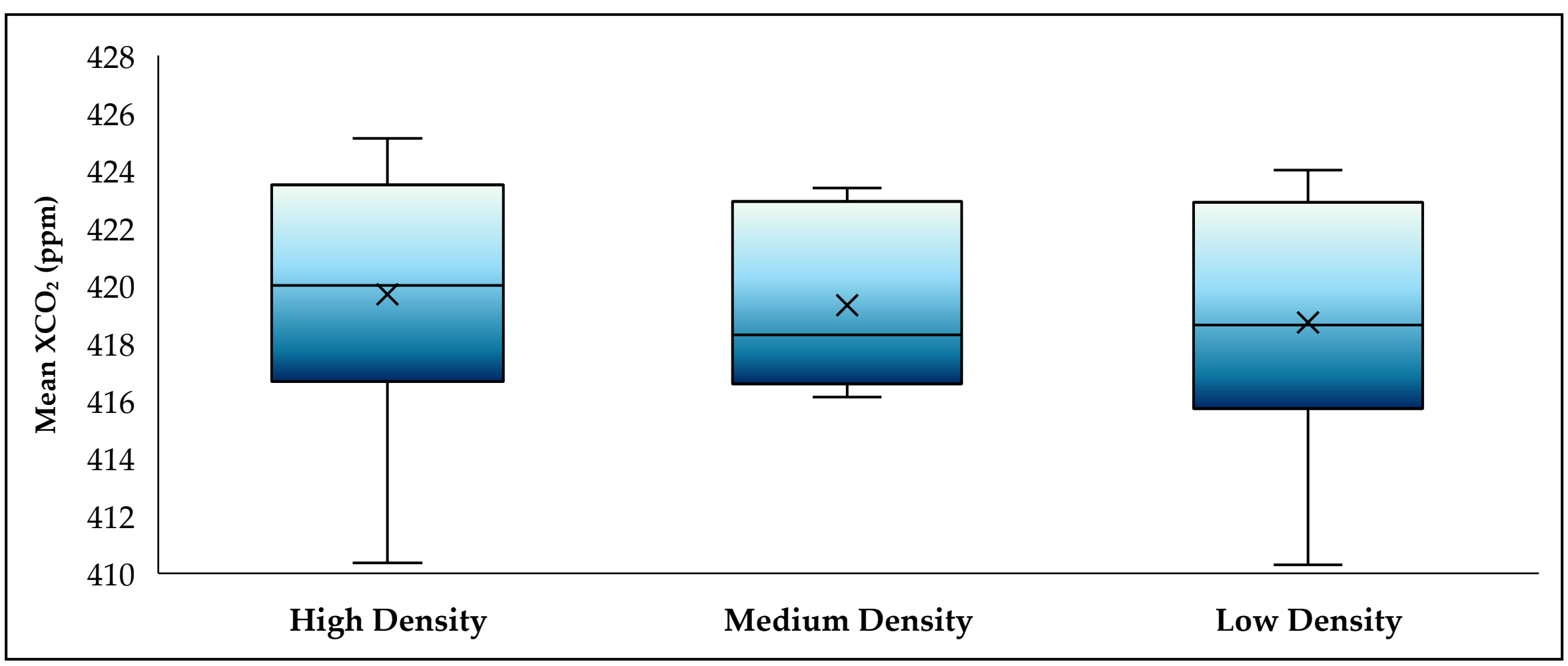

**Fig. 9 Annual Mean $XCO_2$ Distribution by Poultry Density Category**

Analysis of annual $XCO_2$ concentrations across 14 major Canadian poultry-producing regions revealed a moderate positive correlation with facility density ($r = 0.434$, $p < 0.05$), with density accounting for ~19% of the observed variance; the remaining 81% reflected meteorological, topographic, and other environmental influences. A clear density-dependent gradient was evident (Table S6): high-density regions (≥0.0015 facilities/km²) exhibited the highest mean $XCO_2$ (419.15 ± 1.22 ppm), followed by medium-density regions (0.00075–0.00149 facilities/km²; 419.07 ± 1.07 ppm), and low-density regions (<0.00075 facilities/km²; 418.26 ± 1.41 ppm).

Environmental modulation of poultry-driven signals produced notable anomalies (Supplementary File S5). Nova Scotia's Annapolis Valley, despite high density (0.0023 facilities/km²), showed a low mean $XCO_2$ (417.21 ppm) due to maritime wind mixing, whereas Manitoba Eastern exhibited elevated $XCO_2$ (420.42 ppm) despite medium density, likely from reduced mixing in continental interior regions. BC Fraser Valley, with the highest density (0.0176 facilities/km²), also displayed elevated $XCO_2$ (420.74 ppm), consistent with topographic trapping. The corresponding spatial observation metrics are summarized in the.

### 3.3.3 Seasonal Patterns Dynamics of Reconstrued $XCO_2$

Seasonal analysis of $XCO_2$ concentrations across poultry density categories (Table 5) revealed a consistent sinusoidal pattern, with spring maxima (~423 ppm) and summer minima (~415–417 ppm), aligning with Northern Hemisphere $CO_2$ seasonal dynamics driven by vegetation phenology and atmospheric transport. The amplitude of seasonal oscillation varied with production intensity, showing the largest peak-to-trough difference in high-density regions (9.57 ppm), followed by low-density (7.81 ppm) and medium-density (5.87 ppm) categories, indicating a positive association between poultry density and seasonal variability magnitude.

**Table 5. Seasonal $XCO_2$ (ppm)Variations by Poultry Density Category**

| Season | High Density | Medium Density | Low Density |
|---|---|---|---|
| Winter | 425.01 ± 0.10 (n=2) | - | 420.95 ± 0.00 (n=1) |
| Spring | 423.07 ± 0.81 (n=5) | 423.00 ± 0.30 (n=4) | 423.30 ± 0.64 (n=5) |
| Summer | 415.44 ± 3.33 (n=5) | 417.27 ± 0.67 (n=4) | 415.48 ± 1.11 (n=5) |
| Fall | 418.13 ± 1.40 (n=4) | 417.13 ± 1.12 (n=3) | 416.93 ± 3.40 (n=5) |
| **Seasonal Amplitude** | 9.57 | 5.87 | 7.81 |

High-density regions exhibited elevated summer variability ($\sigma = 3.33$ ppm), reflecting non-uniform adjustments likely linked to facility-level factors such as ventilation strategies and sensitivity to heat stress. Medium-density regions showed comparatively low within-season dispersion ($\sigma = 0.67$–1.12 ppm), indicating more stable atmospheric conditions and attenuated anthropogenic influence. Low-density regions, while showing moderate seasonal amplitude, displayed disproportionately high fall variability ($\sigma = 3.40$ ppm), suggesting lagged biophysical or agricultural responses extending beyond summer minima. These findings underscore the interplay between anthropogenic intensity and regional modulation of $XCO_2$ seasonal patterns, highlighting how poultry production density influences both the magnitude and variability of local atmospheric $CO_2$ signals.

### 3.4.4 Monthly Trends and Sub-Seasonal Dynamics

Monthly resolution analysis of reconstrued $XCO_2$ concentrations by poultry density category (Fig. 10) reveals statistically robust seasonal patterns consistent across all categories, with spring-to-fall differences ranging between approximately 7.0 and 7.4 ppm. This uniform amplitude across production intensities indicates synchronized seasonal atmospheric responses likely driven by regional biospheric and agricultural cycles.

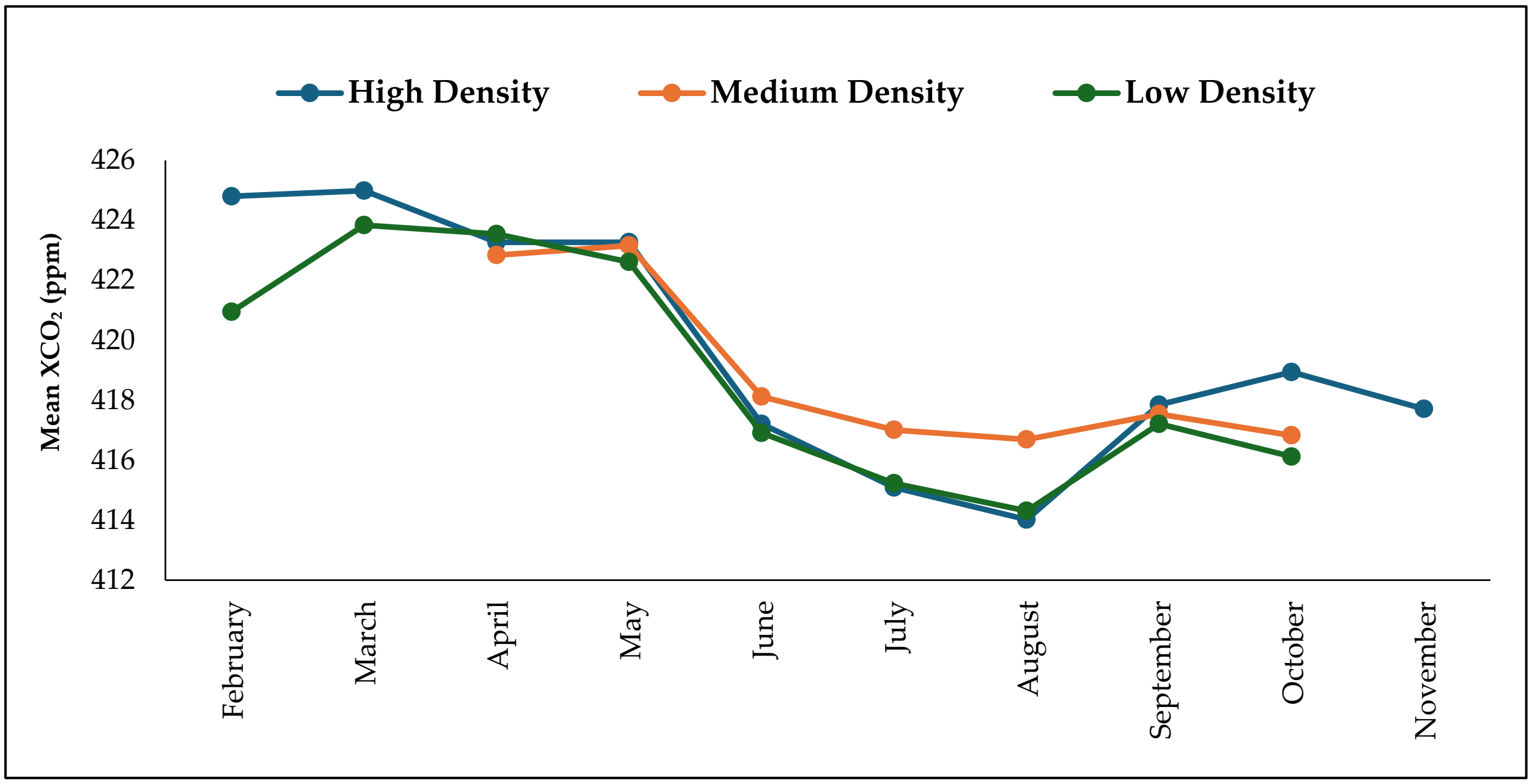

**Fig. 10 Monthly $XCO_2$ Trend by Poultry Density Category**

The March-to-April transition Supplementary File S5 in $XCO_2$ was markedly steeper in high-density regions, with a mean decrease of -1.72 ppm compared to the milder -0.30 ppm change in low-density regions, reflecting greater synchronicity in intensive production schedules, while low-density regions exhibited milder changes. Across all categories, the lowest $XCO_2$ occurred in August, corresponding to peak ecosystem carbon uptake during the growing season. Observation frequency was highest from May through September, while winter months had sparse coverage, limiting data availability. These patterns indicate consistent seasonal dynamics of $XCO_2$ across poultry density categories, with variation in transition rates corresponding to production intensity and temporal resolution.

## 3.5 Cross-Validation with Ground-Based Observations

### 3.5.1 Multi-Scale Spatial Validation

Spatial validation of the reconstructed $XCO_2$ dataset across multiple scales (Table 7) demonstrated excellent agreement with TCCON observations. The 3° radius, representing an intermediate regional scale, showed the strongest performance, with reconstructed $XCO_2$ averaging 418.12 ± 1.78 ppm and minimal bias of –0.14 ppm (–0.03% relative error) relative to the TCCON mean of 418.26 ± 1.41 ppm. RMSE was 0.14 ppm with a normalized RMSE of 0.033%, and a high Kolmogorov-Smirnov p-value of 0.924 indicated statistically indistinguishable distributions, validating the model's uncertainty framework.

**Table 7. Multi-scale Spatial Validation Across Validation Domains**

| Validation Domain | TCCON Mean (ppm) | Reconstructed Mean (ppm) | Bias (ppm) | RMSE (ppm) | NRMSE (%) | KS p-value |
|---|---|---|---|---|---|---|
| 1° radius | 417.89 | 417.34 | -0.55 | 0.55 | 0.132 | 0.857 |
| 3° radius | 418.26 | 418.12 | -0.14 | 0.14 | 0.033 | 0.924 |
| 5° radius | 418.45 | 418.89 | 0.44 | 0.44 | 0.105 | 0.762 |

At the 1° radius, representing local-scale validation, the reconstructed mean of 417.34 ppm produced a bias of –0.55 ppm and RMSE of 0.55 ppm. The standard deviation ratio of 0.92 shows that 92% of local variability is captured, though the small sample size (N = 5) increases uncertainty (95% confidence interval ±0.41 ppm). At the 5° radius, encompassing continental-scale processes, a slight positive bias of +0.44 ppm (+0.11% relative error) was observed, reflecting scale-dependent averaging, sub-grid process sensitivity, and spatial smoothing. Across all domains, absolute biases remained below 0.6 ppm, with a mean spatial MAE of 0.377 ppm. Cohen's d at 3° radius was negligible (0.087), confirming no practically significant difference between reconstructed and observed distributions at regional scales relevant for agricultural emission monitoring.

### 3.5.2 Temporal Validation and Seasonal Cycle Fidelity

Monthly comparisons between TCCON and ST-ViWT demonstrate the model's ability to capture temporal variability and seasonal dynamics. Statistical evaluation (Fig. 11) shows strong fidelity with $r = 0.928$ ($R^2 = 0.860$, $p < 0.001$), RMSE = 0.857 ppm, and mean bias = 0.664 ppm. The model successfully captures the Northern Hemisphere seasonal cycle, including the spring maximum (April: 423.89 vs. 422.28 ppm) and late-summer minimum (August: 414.97 vs. 414.98 ppm), with August bias of -0.01 ppm, highlighting accurate representation of peak growing-season $CO_2$ uptake driven by vegetation photosynthesis.

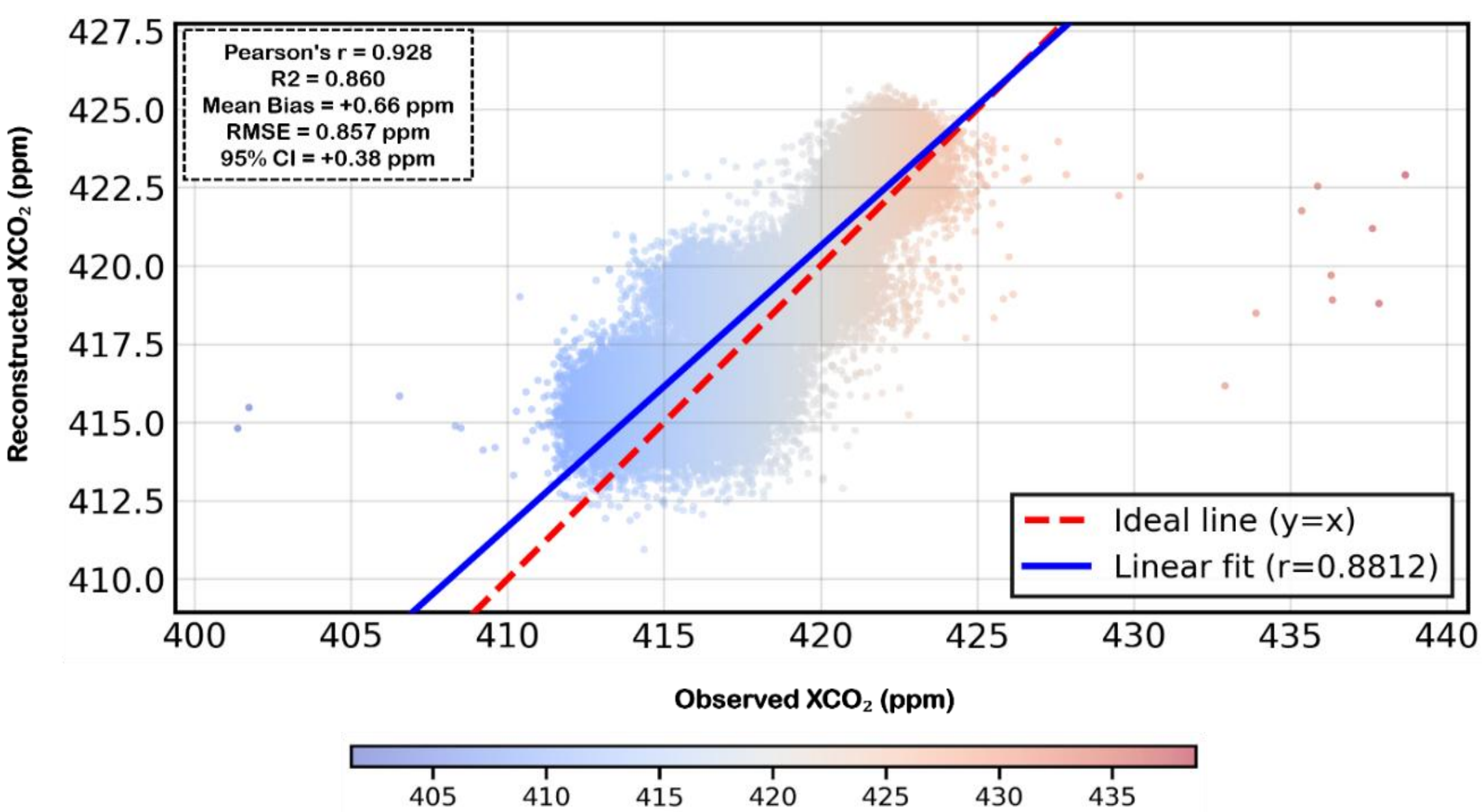

**Fig. 11 Validation of ST-ViWT Reconstructed $XCO_2$ Against TCCON Observed $XCO_2$**

Linear regression of reconstructed $XCO_2$ against TCCON (Reconstructed $XCO_2$ = 1.027 × TCCON – 11.57) indicates slight temporal amplitude amplification, with a 2.7% overestimation. Monthly biases are predominantly positive, averaging +0.66 ± 0.38 ppm (+0.16% relative bias), with maximum overestimation during spring and transitional months (March: +1.06 ppm; April: +1.61 ppm; July: +1.51 ppm). Late summer and early autumn (August–September) show near-zero or slightly negative biases, accurately capturing minimum $CO_2$ concentrations. These results suggest that amplitude amplification primarily affects the spring maximum and seasonal transitions, while summer uptake and baseline levels remain well represented.

## 4. Discussion

### 4.1 ST-ViWT Model Performance and Emission Dynamics

The ST-ViWT framework demonstrated exceptional predictive performance for $XCO_2$ reconstruction over Canadian poultry-intensive regions, achieving validation $R^2 = 0.984$ and RMSE = 0.468 ppm. This represents significant advancement compared to traditional machine learning approaches, with error rates substantially lower than the 0.58-0.91 ppm RMSE reported for XGBoost-based downscaling methods (He et al., 2024; Alcibahy et al., 2025). The model maintained prediction errors within ±1 ppm for 92.3% of observations, demonstrating practical utility for high-precision atmospheric monitoring applications (Kulawik et al., 2019). Superior performance can be attributed to several architectural innovations. The integration of wavelet-based time-frequency analysis addresses fundamental limitations of traditional Fourier transforms in capturing non-stationary temporal patterns characteristic of agricultural $CO_2$ emissions (Jung et al., 2020; Reichert et al., 2024). The geographic positional encoding strategy, incorporating latitude and longitude dependencies, enabled the model to learn region-specific atmospheric transport patterns (Hu et al., 2024). While poultry facility density explained 19% of $XCO_2$ variance, the remaining 81% reflects complex interactions between meteorological, topographic, and ecological factors successfully integrated through our multi-modal framework. The observed seasonal $XCO_2$ amplitude of 8.92 ppm closely with established Northern Hemisphere carbon cycle dynamics (Bastos et al., 2020), confirming accurate representation of large-scale biogeochemical processes. However, differential seasonal responses across poultry density categories reveal important insights. High-density regions exhibited the largest seasonal amplitude (9.57 ppm), compared to medium-density (5.87 ppm) and low-density (7.81 ppm) regions, suggesting that intensive poultry production may amplify natural seasonal variations through multiple pathways including feed crop production and energy-intensive housing systems (Grossi et al., 2019; Cheng et al., 2022).

### 4.2 Attribution Challenges: Confounding Sources and Biospheric Masking

The moderate correlation between poultry facility density and $XCO_2$ concentrations ($r = 0.434$) should be interpreted with caution due to confounding from co-located agricultural activities. High-density poultry regions frequently coincide with intensive crop production, particularly feed grains such as corn, wheat, and soybeans, which contribute substantial $CO_2$ fluxes through soil heterotrophic respiration, decomposition of crop residues, and fossil fuel use from farm machinery (Bastos et al., 2020). Agricultural soils can emit 2-8 t $CO_2$-C $ha^{-1}$ $yr^{-1}$ via heterotrophic respiration alone (Liang et al., 2024). In comparison, a typical high-density poultry facility consuming 500 MWh $yr^{-1}$ produces 150-200 t $CO_2$ annually, whereas intensive maize cultivation on 1,000 ha may release 2,000-8,000 t $CO_2$, representing a 10-40-fold difference (Huang and Guo, 2018). This disparity suggests that the observed density of $XCO_2$ correlation likely reflects the broader agricultural landscape rather than direct poultry emissions. The pronounced summer minimum in $XCO_2$ (414-416 ppm in August) across all density categories demonstrates the dominant influence of biospheric uptake, which masks anthropogenic differences during peak growing season (Bastos et al., 2020). Canadian croplands exhibit gross primary productivity of 4-12 g C $m^{-2}$ $day^{-1}$, corresponding to 15-45 t $CO_2$-C $km^{-2}$ $day^{-1}$, far exceeding typical poultry emissions of 0.4–2.7 t $CO_2$ $day^{-1}$ per facility (Ashton et al., 2023). This is reflected in the reduced summer variability across density categories ($\sigma = 0.64$ ppm) compared to spring ($\sigma = 1.21$ ppm), when limited photosynthetic activity allows emission heterogeneity to be more clearly observed (Kivimäki et al., 2025). Regional anomalies further illustrate the role of local meteorology and topography. The Nova Scotia Annapolis Valley, despite high facility density (0.0023 facilities $km^{-2}$), recorded lower $XCO_2$ (417.21 ppm) due to maritime mixing, whereas the BC Fraser Valley exhibited elevated concentrations (420.74 ppm) at the highest density (0.0176 facilities $km^{-2}$), likely due to topographic trapping (Jacobs et al., 2024).

### 4.3 Model Validation and Temporal Fidelity

Multi-scale spatial validation against East Trout Lake TCCON provided critical independent verification of ST-ViWT reconstruction accuracy beyond satellite-derived training data (Mostafavi Pak et al., 2023). At the optimal 3° radius, the model showed exceptional agreement with TCCON (bias = -0.14 ppm, RMSE = 0.140 ppm, NRMSE = 0.033%), outperforming typical satellite validation requirements of 0.2–0.25 ppm bias (Laughner et al., 2024; Das et al., 2025). Scale-dependent biases, ranging from -0.55 ppm at 1° to +0.44 ppm at 5°, highlight systematic representation effects likely driven by error averaging and sensitivity to subgrid transport processes (Gadhavi et al., 2025). Temporal validation across eleven months confirmed strong seasonal fidelity ($r = 0.928$, $R^2 = 0.860$, $p < 0.001$), capturing the spring maximum and late-summer minimum typical of Northern Hemisphere dynamics (Bastos et al., 2020; Ruehr et al., 2023). August agreement was near-perfect (bias = -0.01 ppm), while modest spring overestimation (April: +1.61 ppm) suggests slight amplitude amplification from synthetic gap-filling. With ~0.25% precision (~1 ppm), the TCCON standard underpins global GHG validation (Laughner et al., 2024), reinforcing the robustness of these results. This framework supports operational agricultural emission monitoring, though limited TCCON coverage in poultry-intensive regions highlights the need for expanded ground-based networks (Mostafavi Pak et al., 2023).

### 4.4 Implications for Climate Monitoring and Research Directions

For Canadian agricultural GHG reporting, the framework provides a pathway for enhancing National Inventory Report accuracy through atmospheric verification of localized agricultural emission estimates. Current methodologies rely predominantly on IPCC Tier 1 and Tier 2 approaches using activity data and emission factors, which often fail to capture the spatial heterogeneity and temporal dynamics of intensive livestock regions. Satellite-based $XCO_2$ monitoring provides independent constraints on bottom-up estimates and supports the development of region-specific emission factors (Watanabe et al., 2023). When integrated with precision livestock technologies, these atmospheric data enable farm-level benchmarking, facilitating producer participation in carbon offset programs.

Future advancements, including higher-resolution satellites, extended temporal coverage, and machine learning based facility detection, can enhance spatial and temporal resolution of emission estimates. Current 0.25° data are biased toward summer and may miss small-scale poultry operations, but integration with airborne measurements and high-resolution imagery can improve spatial and temporal constraints (Robinson et al., 2022). To improve source attribution, integrating multi-tracer satellite observations of $CH_4$ and $NH_3$ with high-resolution land-use, solar-induced fluorescence and crop productivity data, alongside 1-4 km atmospheric inverse modeling, can discriminate poultry emissions from other agricultural sources with minimal biospheric interference (Miller and Michalak, 2020).

## 5. Conclusions

Transforming Earth observation into actionable intelligence for agricultural emissions requires models that can reconcile the discontinuities of satellite sensing with the complexity of real atmospheric dynamics. The Spatiotemporal Vision Transformer with Wavelets (ST-ViWT) framework accomplishes this synthesis. By coupling time–frequency decomposition with transformer attention, it achieves near-benchmark accuracy ($R^2 = 0.984$; RMSE = 0.468 ppm) in reconstructing atmospheric $CO_2$ ($XCO_2$) across Canada's poultry-intensive regions, while recovering over 92 % of missing satellite data with quantified uncertainty. Independent comparison with TCCON ground observations confirms the model's reliability with spatial bias as low as −0.14 ppm and temporal coherence ($r = 0.928$), capturing the true amplitude of seasonal oscillations in agricultural carbon flux. The spatial gradients revealed, including a moderate density-dependent enhancement ($r = 0.43$) and distinct 9.57 ppm seasonal amplitude in high-intensity production zones, expose previously unresolved heterogeneity in Canada's livestock-linked carbon landscape. Beyond accuracy, ST-ViWT represents a conceptual advance: a generalizable GeoAI paradigm capable of integrating multi-modal satellite and surface data to deliver continuous, high-fidelity atmospheric fields. Its scalability positions it as a foundation model for environmental monitoring, supporting transparent national carbon accounting and adaptive management of agro-ecosystems. Future extensions including linking trace gases and canopy fluorescence will enable attribution of sector specific fluxes at sub-kilometer precision, forging a new standard for AI-driven climate observatories that bridge atmospheric science, remote sensing, and sustainable food production.

## Supplementary Data

1. Supplementary File S1: Distribution of Poultry across Regions.
2. Supplementary File S2: Feature Engineering and Model Formulations.
3. Supplementary File S3: Model Performance Metrics.
4. Supplementary File S4: Spatial Analysis Results.
5. Supplementary File S4: Temporal Analysis Results.

## Data Availability Statement

All datasets, preprocessing scripts, model configurations, and supplementary materials are openly available at https://github.com/MooAnalytica/geoai-xco2-canada, with the Canadian poultry facility dataset accessible upon reasonable request due to confidentiality.

## Funding

This work is kindly sponsored by the Natural Sciences and Engineering Research Council of Canada (RGPIN 2024-04450), the Net Zero Atlantic Canada Agency (300700018), Mitacs Canada (IT36514), and the Department of New Brunswick Agriculture, Aquaculture and Fisheries (NB2425-0025).